[1]D. Dhinakaran
[2]S. Edwin Raja
[3]M. Thiyagarajan
[4]J. Jeno Jasmine
[5]P. Raghavan

# Optimizing Disease Prediction with Artificial Intelligence Driven Feature Selection and Attention Networks

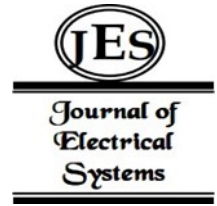

*Abstract:* - The rapid integration of machine learning methodologies in healthcare has ignited innovative strategies for disease prediction, particularly with the vast repositories of Electronic Health Records (EHR) data. This article delves into the realm of multi-disease prediction, presenting a comprehensive study that introduces a pioneering ensemble feature selection model. This model, designed to optimize learning systems, combines statistical, deep, and optimally selected features through the innovative Stabilized Energy Valley Optimization with Enhanced Bounds (SEV-EB) algorithm. The objective is to achieve unparalleled accuracy and stability in predicting various disorders. This work proposes an advanced ensemble model that synergistically integrates statistical, deep, and optimally selected features. This combination aims to enhance the predictive power of the model by capturing diverse aspects of the health data. At the heart of the proposed model lies the SEV-EB algorithm, a novel approach to optimal feature selection. The algorithm introduces enhanced bounds and stabilization techniques, contributing to the robustness and accuracy of the overall prediction model. To further elevate the predictive capabilities, an HSC-AttentionNet is introduced. This network architecture combines deep temporal convolution capabilities with LSTM, allowing the model to capture both short-term patterns and long-term dependencies in health data. Rigorous evaluations showcase the remarkable performance of the proposed model. Achieving a 95% accuracy and 94% F1-score in predicting various disorders, the model surpasses traditional methods, signifying a significant advancement in disease prediction accuracy. The implications of this research extend beyond the confines of academia. By harnessing the wealth of information embedded in EHR data, the proposed model presents a paradigm shift in healthcare interventions. The optimized diagnosis and treatment pathways facilitated by this approach hold promise for more accurate and personalized healthcare, potentially revolutionizing patient outcomes

*Keywords:* Multi-disease prediction, SEV-EB, Ensemble feature selection, HSC-AttentionNet, Machine learning, Healthcare, EHR data

## I. INTRODUCTION

The advent of machine learning in healthcare has ushered in a transformative era, providing unprecedented opportunities to revolutionize disease prediction and patient care [1]. The integration of advanced technologies, particularly in the domain of predictive analytics, holds the potential to redefine diagnostic capabilities and treatment strategies. In this context, our research embarks on a journey to explore and enhance multi-disease prediction, a critical aspect of proactive healthcare management. The escalating volume of healthcare data, particularly Electronic Health Records (EHR), has emerged as a goldmine for predictive modeling and decision support systems [2]. The rich tapestry of patient information encapsulated in EHRs offers a comprehensive view of an individual's health journey, presenting an opportunity to predict the onset of various diseases before they manifest clinically [3]. Traditional healthcare models, often constrained by the limitations of manual analysis and heuristic approaches, pale in comparison to the potential that machine learning algorithms bring to the table.

While individual disease prediction models have made notable strides, a crucial gap persists in the realm of multi-disease prediction. The complexity of the human body and the interconnectedness of various health parameters necessitate a holistic approach that transcends the siloed nature of single-disease prediction. Addressing this gap becomes imperative as healthcare professionals increasingly encounter patients with overlapping or co-existing medical conditions [4]. The ability to predict and manage multiple diseases concurrently holds profound implications for personalized treatment plans and proactive healthcare interventions.

However, the journey towards effective multi-disease prediction is not devoid of challenges. The sheer diversity and heterogeneity of healthcare data, coupled with the need for interpretability in clinical settings, demand sophisticated and robust models [5]. Furthermore, the dynamic nature of health conditions, the evolution of diseases over time, and the need for real-time predictions pose additional hurdles. In response to

[1,2,3]Department of Computer Science and Engineering, Vel Tech Rangarajan Dr. Sagunthala R&D Institute of Science and Technology, Chennai, India
[4]Department of Computer Science and Engineering, R.M.K. Engineering College, Tamil Nadu, India
[5]Department of Computer Science and Engineering, P.S.R. Engineering College, Sivakasi, India
drdhinakarand@veltech.edu.in, edwinrajas@gmail.com, thiyaga1647@gmail.com, jenojasmine@gmail.com, raghavan.ramesh1988@gmail.com

these challenges, our research endeavors to contribute to the evolving landscape of multi-disease prediction [6-8]. We propose an innovative ensemble approach, leveraging the synergy of statistical features, deep features, and optimally selected features. At the heart of our model lies the Stabilized Energy Valley Optimization with Enhanced Bounds (SEV-EB) algorithm, a novel methodology designed to optimize feature selection while enhancing stability and accuracy.

Our primary objective is to develop a predictive model that not only excels in accuracy but also demonstrates adaptability to the dynamic nature of healthcare data. By integrating the SEV-EB algorithm into our ensemble model, we aim to achieve a harmonious balance between predictive power and stability. Additionally, we aspire to showcase the effectiveness of our approach in predicting a spectrum of diseases, thereby contributing to the realization of comprehensive and proactive healthcare strategies. This research holds immense significance in the context of advancing healthcare practices. Successful implementation of our proposed model could empower healthcare professionals with timely and accurate insights, allowing for early interventions and personalized treatment plans. The potential impact extends beyond individual patient care to broader public health initiatives, where early identification of disease trends could inform preventive measures and healthcare resource allocation.

*1.1 Research gaps and challenges*

As we embark on the journey to advance multi-disease prediction using machine learning in healthcare, it is essential to acknowledge the existing research gaps and challenges that underscore the complexity of this field. Identifying and addressing these gaps is crucial for developing robust models and ensuring the successful integration of predictive analytics into clinical practice.

*Lack of Comprehensive Datasets:*

The availability of comprehensive and diverse datasets that capture the intricacies of multi-disease scenarios is often limited [9]. Many existing datasets predominantly focus on individual diseases, overlooking the complexities that arise when multiple conditions coexist. Curating and standardizing datasets that encompass a broad spectrum of health parameters, demographics, and multi-morbidity scenarios is a formidable challenge. The development of such datasets requires collaboration between healthcare institutions, data scientists, and regulatory bodies.

*Interconnectedness of Diseases:*

Understanding the intricate relationships and interconnectedness between different diseases remains a significant research gap. Existing models often treat diseases in isolation, neglecting the potential synergies or antagonisms that may influence the predictive accuracy of multi-disease models [10-12]. Developing models that account for the complex interplay between various diseases, considering how the presence of one condition might impact the likelihood or progression of another, poses a challenge. Integrating this nuanced understanding into machine learning models requires sophisticated algorithmic approaches.

*Real-time Predictions and Adaptability:*

Many existing models struggle to provide real-time predictions, particularly in dynamic healthcare environments where patient conditions may rapidly evolve [13]. There is a gap in the adaptability of models to changing health statuses and the need for timely interventions. Achieving real-time predictions demands not only algorithmic sophistication but also considerations for the computational infrastructure required. Developing models that can adapt to changing health conditions dynamically is a challenge that requires continuous refinement and optimization.

*Model Interpretability in Clinical Settings:*

The interpretability of machine learning models in clinical settings remains a substantial gap. Healthcare professionals often require models to provide interpretable insights to build trust and facilitate informed decision-making [14]. Balancing the complexity of advanced machine learning models with the need for interpretability in clinical settings is challenging. Developing models that can provide actionable insights while being transparent and interpretable is an ongoing challenge.

*Ethical and Privacy Concerns:*

The ethical implications and privacy concerns associated with the use of patient data for predictive analytics pose a substantial gap in current research. Ensuring the responsible and ethical use of sensitive healthcare information is a critical aspect that requires careful consideration. Striking the right balance between leveraging the power of data for predictive modeling and safeguarding patient privacy is an ongoing challenge. Developing robust frameworks that adhere to ethical guidelines and regulatory standards is essential.

*Generalization to Diverse Populations:*

Many machine learning models struggle with generalizing predictions across diverse populations. There is often a gap in understanding how well models trained on specific datasets perform when applied to populations with different demographics, healthcare systems, or socioeconomic contexts [15]. Achieving robust generalization requires careful consideration of biases in training data, model bias, and external validation on diverse datasets. Addressing these challenges is crucial for ensuring the inclusivity and fairness of predictive models.

Addressing these research gaps as well as the challenges requires collaborative efforts from researchers, healthcare practitioners, data scientists, and policymakers. By systematically addressing these gaps, the field of multi-disease prediction can move towards more effective, ethical, and practical applications in healthcare. In the subsequent sections of this research article, we delve into the methodology, experimental design, results, and implications of our multi-disease prediction model. By sharing our insights, findings, and reflections, we aim to contribute substantively to the ongoing discourse on the application of ML in healthcare and, more specifically, in the realm of multi-disease prediction.

## II. RELATED WORKS

The following literature survey provides an overview of existing research related to the domain of our work. While the field is broad and dynamic, our focus remains on key aspects that directly influence and contribute to the context of our study. A significant body of literature explores the application of various machine learning models in classification tasks. These models include neural networks, decision trees, random forests, and ensemble methods. Studies emphasize the importance of selecting appropriate models based on the characteristics of the dataset and the nature of the classification problem. Comparative analyses of different machine learning models provide valuable insights into their strengths and weaknesses. Studies in this domain often consider the performance metrics mentioned above, offering a nuanced understanding of how various models perform under different conditions.

In the study conducted by Babukarthik et al. [16], a novel method is presented for the identification of COVID-19 pneumonia and healthy lung conditions within chest X-ray (CXR) images. The researchers employed deep learning techniques, specifically utilizing the Genetic Deep Learning Convolutional Neural Network (GDCNN), a cutting-edge approach in the field. GDCNN was trained from the ground up to extract distinctive features that differentiate COVID-19 cases from normal images. The investigation involved a comprehensive dataset consisting of over 5,000 CXR images, covering instances of pneumonia, normal cases, and other pneumonia types. Noteworthy is the observation that training GDCNN from scratch yielded superior performance compared to established transfer learning methodologies commonly employed for this specific task.

Shreshth et al. [17] introduce an enhanced mathematical model designed for the analysis and prediction of the COVID-19 epidemic's growth. The study employs a machine learning-based approach to anticipate potential threats across global regions. Through the application of iterative weighting to the Generalized Inverse Weibull distribution, the authors demonstrate a more effective fitting method for constructing a predictive framework. This framework is subsequently deployed on a cloud computing platform, enhancing its capabilities for generating more accurate and real-time predictions regarding the epidemic's growth patterns. The authors underscore the significance of data-driven approaches with heightened accuracy, emphasizing their role in enabling proactive responses from both governmental entities and citizens. The study concludes by suggesting potential avenues for further research and establishing the groundwork for practical applications of their model.

In their pursuit to enhance triage procedures during the COVID-19 pandemic, Zoabi et al. [18] developed a machine-learning model focused on predicting test results. The training phase of the model involved 51,831 individuals with recorded test outcomes, encompassing 4,769 confirmed cases. Further evaluation was conducted on a separate test set comprising 47,401 individuals, with 3,624 confirmed cases. The model, leveraging a modest set of eight binary features, such as sex, age, contact history with an infected individual, and five initial symptoms, exhibited notable accuracy in predicting COVID-19. Drawing on nationwide data from the Israeli Ministry of Health, this study underscores the model's potential for identifying cases based on easily obtainable information. The proposed framework serves as a valuable tool for prioritizing testing efforts and optimizing resource allocation, especially in environments with constrained healthcare infrastructure.

Sanzida et al. [19] investigate automatic COVID-19 detection using machine learning techniques, aiming to construct an intelligent web application. The study involves rigorous dataset preprocessing, including null value elimination, feature engineering, and synthetic oversampling techniques (SMOTE). It evaluates a comprehensive range of classifiers, spanning logistic regression, random forest, decision tree, k-nearest neighbor, support vector machine (SVM), ensemble models (adaptive boosting and extreme gradient boosting), and deep learning techniques (artificial neural network, convolutional neural network, and long short-term memory). To enhance prediction interpretability, Explainable AI is employed, specifically the LIME framework. The study identifies a hybrid CNN-LSTM algorithm paired with SMOTE as the most successful model, achieving 96.34% accuracy and a 0.98 F1 score on an open-source dataset from the Israeli Ministry of Health. This model is subsequently deployed on a website, enabling users to receive immediate COVID-19 prognoses based on their symptoms, thus demonstrating its practical application.

Painuli et al. [20] present methodologies for forecasting future COVID-19 cases based on existing data, employing machine learning approaches. The study delves into two distinct solutions: one for predicting the likelihood of infection and another for forecasting the actual number of positive cases. The authors conducted trials involving various algorithms, with emphasis placed on the algorithm exhibiting the highest accuracy. The chapter specifically explores the application of the autoregressive integrated moving average time series for forecasting confirmed cases across different states in India. Among the classifiers considered, namely random forest and extra tree classifiers, both demonstrated impressive accuracies exceeding 90%. Notably, the extra tree classifier outperformed with an accuracy of 93.62%. These results provide valuable insights for governmental bodies, enabling them to implement proactive measures in response to the identified risk levels. The availability of accurate forecasting techniques serves as a crucial tool in the collective effort to combat the COVID-19 pandemic.

Alanazi et al. [21] conducted a study with the objective of applying computational methods, specifically machine learning techniques, to predict stroke based on lab test data. The researchers utilized datasets from the National Health and Nutrition Examination Survey, employing three distinct data selection methods: without data resampling, with data imputation, and with data resampling. The development of predictive models involved the use of four machine learning classifiers, and the performance of these models was assessed using six performance measures. Tahia et al. [22] employed a variety of physiological parameters and machine learning algorithms, including Logistic Regression (LR), Decision Tree (DT) Classification, Random Forest (RF) Classification, and Voting Classifier, to train four distinct models for reliable stroke prediction. Among these algorithms, Random Forest emerged as the most effective, achieving an accuracy of approximately 96 percent. The researchers utilized the open-access Stroke Prediction dataset for model development.

Notably, the accuracy percentages obtained in this investigation surpassed those reported in previous studies, indicating the heightened reliability of the models employed. Extensive model comparisons were conducted to establish their robustness, and the study's analysis provides a clear insight into the effectiveness of the proposed scheme for stroke prediction. In the realm of machine learning and classification, our work delves into a dynamic field characterized by diverse methodologies and evolving techniques. Central to our study is the exploration of various machine learning models, including neural networks, decision trees, and ensemble methods, each selected based on the intricacies of the dataset and the specific classification challenge at hand. A pivotal aspect of our research revolves around the strategic choice of activation functions in neural networks, considering the impact of functions like ReLU, Sigmoid, Tanh, Leaky ReLU, and Softmax on convergence speed and overall accuracy.

Recognizing the significance of robust model evaluation, our study closely aligns with the prevailing literature, emphasizing the use of comprehensive metrics such as accuracy, precision, recall, F1 Score, and AUC-ROC. Ensemble learning methods, involving the combination of multiple models, have garnered attention for their capacity to enhance accuracy and resilience in classification tasks, aligning with our exploration of optimized machine learning models. Our work also contributes to the ongoing discourse on optimization techniques, shedding light on the importance of hyperparameter tuning and advanced algorithms to refine model performance. Furthermore, our research situates itself within the framework of comparative model analysis, drawing insights from existing studies that evaluate the strengths and weaknesses of different machine learning approaches. In addressing the complexities of the domain, our study tackles a specific problem with a focus on providing innovative solutions. By leveraging the insights from the literature survey, we aim to contribute novel perspectives and advancements, positioning our research at the forefront of current developments in machine learning and classification methodologies.

III. PROPOSED MODEL

*3.1 Motivation*

The motivation behind this research stems from the profound impact that accurate and timely disease prediction can have on the landscape of healthcare. As we confront the complexities of modern healthcare systems, characterized by a surge in data volume and the increasing prevalence of multi-morbidity scenarios, the need for advanced predictive models becomes more evident than ever. Our motivation is grounded in addressing the following key factors:

*Proactive Healthcare Management:*

The traditional paradigm of reactive healthcare is evolving into a proactive model that emphasizes prevention and early intervention [23]. Accurate disease prediction models empower healthcare professionals to identify potential health risks before they manifest clinically, paving the way for proactive and personalized healthcare strategies.

*Rising Healthcare Burden:*

The escalating burden of chronic and complex diseases poses a significant challenge to healthcare systems globally. Multi-disease scenarios, where individuals contend with multiple health conditions simultaneously, necessitate a paradigm shift in disease prediction models [24]. Our motivation is to contribute to the development of solutions that can effectively navigate this intricate healthcare landscape.

*Potential for Improved Patient Outcomes:*

Early detection of diseases translates into timely interventions and tailored treatment plans, which have the potential to significantly improve patient outcomes. By predicting multiple diseases concurrently, our research aspires to empower healthcare providers with actionable insights that lead to more effective and holistic patient care.

*Harnessing the Power of Machine Learning:*

The advancements in machine learning present an unprecedented opportunity to extract meaningful patterns and insights from vast and complex healthcare datasets. Motivated by the transformative potential of these technologies, our research aims to harness the power of ML to enhance the accuracy as well as the efficiency of disease prediction models.

*Bridging Gaps in Current Research:*

The motivation also arises from identifying and addressing critical gaps in existing research. We seek to contribute to the evolving field of multi-disease prediction by proposing innovative methodologies, such as the Stabilized Energy Valley Optimization with Enhanced Bounds (SEV-EB) algorithm, that address the challenges of interconnected diseases and real-time adaptability [25].

*Impact on Public Health:*

The broader societal impact is a key driver of our research motivation. By improving the accuracy of disease predictions and optimizing healthcare pathways, our work aspires to contribute to public health initiatives [26-28]. This includes early identification of disease trends, resource optimization, and the potential to mitigate the long-term societal and economic impact of prevalent health conditions.

Our motivation is deeply rooted in the belief that leveraging cutting-edge technologies to predict and manage diseases proactively can reshape the future of healthcare. By aligning our research goals with the evolving needs of healthcare systems and the well-being of individuals, we aim to make tangible contributions that transcend the boundaries of traditional healthcare approaches.

*3.2 Key Contributions:*

*Ensemble Feature Selection:* A novel model combining statistical, deep, and optimally selected features, maximizing predictive power.

*Stabilized Energy Valley Optimization (SEV-EB):* Introduction of a new algorithm for optimal feature selection, enhancing stability and accuracy.

*HSC-AttentionNet:* Integration of deep temporal and long-term memory capabilities for effective disease prediction.

*State-of-the-Art Performance:* Significant improvement in accuracy and F1-score compared to existing methods.

*3.3 Multi-Disease Prediction Model*

Within the burgeoning realm of precision medicine, multi-disease prediction systems have emerged as promising tools for proactive healthcare. The Fig. 1 offers a captivating window into the intricate inner workings of such a system, illuminating the journey from data acquisition to risk assessment.

*Step 1: Data Acquisition - Fueling the Predictive Engine*

The system's foundation lies in the input data, encompassing both statistical features and deep features. Statistical features, akin to meticulously recorded measurements, capture vital health parameters through quantitative descriptors like mean, standard deviation, and correlation coefficients. Deep features, however, delving deeper, leverage the prowess of convolutional neural networks (CNNs) to unearth hidden patterns and temporal dynamics embedded within the data, akin to deciphering subtle nuances in medical images or biosignals.

*Step 2: Feature Selection - Extracting the Quintessential Clues*

Not all data points hold equal weight in the pursuit of accurate predictions. SEV-EB, a Stabilized Energy Valley Optimization technique, acts as the discerning detective, meticulously sifting through the data to identify the most informative and predictive features. This process, akin to prioritizing the most critical evidence in a forensic investigation, ensures the model focuses its resources on the most relevant information.

*Step 3: Model Construction - Weaving the Predictive Tapestry*

The heart of the system lies in its prediction model, a carefully crafted ensemble of three powerful components:

*HSC-Attention Net:* This component functions as a vigilant spotlight, directing the model's attention towards the most salient features within the data, similar to how a skilled investigator focuses on key details at a crime scene.

*Deep Temporal Convolutional Network (DTCN):* Akin to a seasoned detective tracing the timeline of events, the DTCN meticulously analyzes the temporal relationships between features, discerning evolving patterns and trends within the data over time.

*Long Short-Term Memory (LSTM):* This component possesses the remarkable ability to retain long-term dependencies within the data, resembling a detective with an exceptional memory, capable of connecting seemingly disparate clues to paint a holistic picture.

*Step 4: Risk Assessment - Unveiling the Probabilistic Landscape*

Having gleaned insights from the data, the model culminates in the generation of predicted outcomes for each disease under investigation [29]. These outcomes, unlike definitive pronouncements, are expressed as probabilities, quantifying the individual's susceptibility to each condition. This probabilistic approach, analogous to the nuanced conclusions drawn by a detective after careful analysis, provides valuable information for personalized healthcare interventions.

*Step 5: Optimization - Refining the Predictive Lens*

The system's pursuit of enhanced accuracy is a continuous endeavor. The SEV-EB technique is employed once again, not only for initial feature selection but also for ongoing model optimization. This iterative process, akin to a detective continuously revisiting and refining their investigative approaches, ensures the system's predictive capabilities remain constantly honed and evolving.

A range of sources, including wearable sensors, medical imaging, and e-health records, can provide the input data. A number of ailments, such as diabetes, heart disease, as well as cancer, can be predicted by the method. The system is still under development, but it has shown promising results in early studies.

*This multi-disease prediction system boasts several advantages:*

*Early detection:* Proactive identification of potential illnesses enables timely interventions and improved patient outcomes.

*Personalized healthcare:* By analyzing individual medical histories, the system can tailor predictions and treatment plans to specific patients.

*Efficiency:* It streamlines diagnosis by predicting multiple diseases simultaneously, eliminating the need for separate tests and consultations.

This multi-disease prediction system paves the way for a future of proactive healthcare, empowering individuals with personalized risk assessments and enabling clinicians to tailor interventions with greater precision. The intricate interplay of data acquisition, feature selection, model construction, and risk assessment, aptly captured in the presented diagram, serves as a testament to the ongoing advancements in medical technology. As research in this domain continues to unfold, multi-disease prediction systems hold immense potential to revolutionize personalized healthcare, ushering in an era of proactive health management and enhanced well-being.

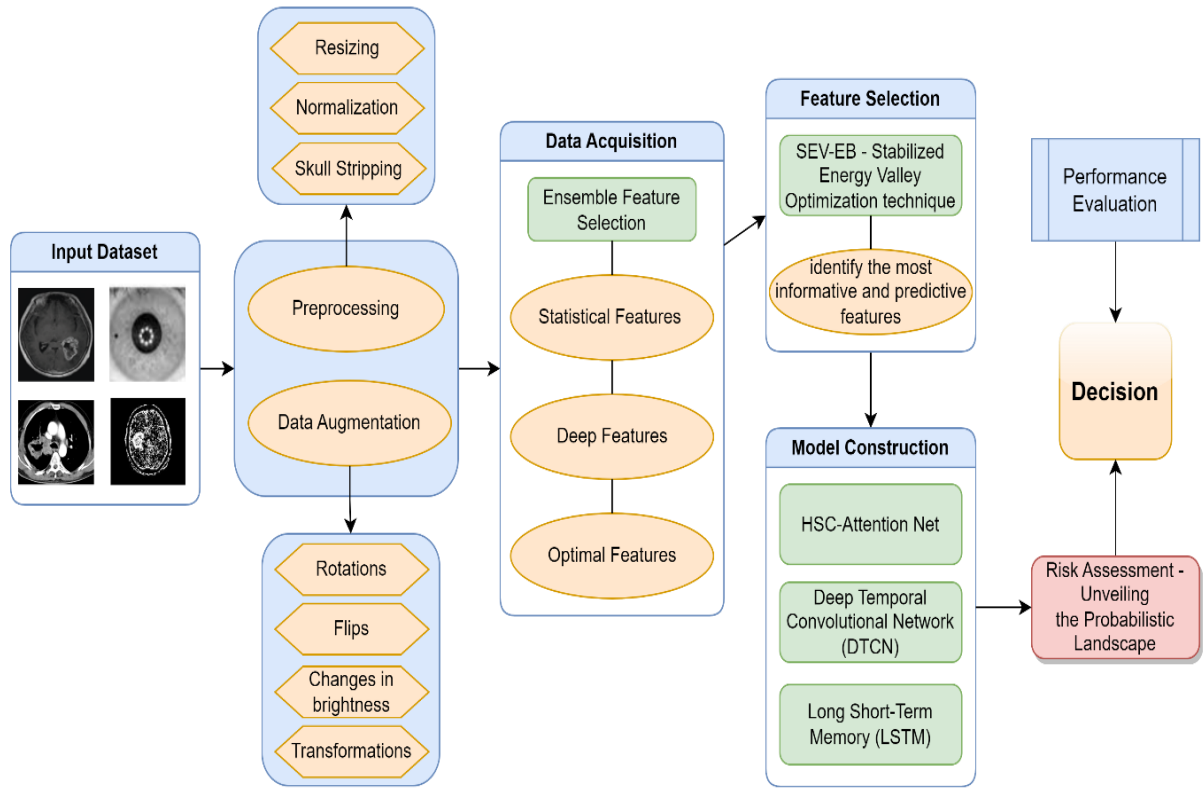

Fig. 1: Multi-Disease Prediction Model - Framework

*3.4 Process Flow:*

Figure 2 outlines a basic flow for the Stabilized Energy Valley Optimization with Enhanced Bounds (SEV-EB) algorithm. Let's break down the flow into steps:

*1. Initialization of Population:*

The initialization phase is a crucial starting point for any optimization algorithm. In the context of SEV-EB, a population is formed, representing potential solutions to the optimization problem. This population is typically generated randomly or based on certain heuristics, creating a diverse set of candidate solutions.

*2. Evaluation of Fitness Value:*

Every person's fitness is assessed after the population has been initiated. The fitness function measures each solution's effectiveness in relation to the optimization objective. This step involves assessing the quality of the solutions, providing a basis for comparison and selection.

*3. Parameter Evaluation:*

In many optimization algorithms, parameters play a significant role in shaping the behavior of the search process. The SEV-EB algorithm involves evaluating these parameters, which could be algorithm-specific or related to the characteristics of the optimization problem itself. Parameter tuning is crucial for fine-tuning the algorithm's performance.

*4. Check for a Condition (i>= 0):*

The algorithm includes a conditional statement, checking whether a variable 'i' is greater than or equal to zero. This condition likely serves as a control mechanism, guiding the algorithm's flow based on the current state of the optimization process. If 'i' is ≥ zero, the algorithm proceeds;

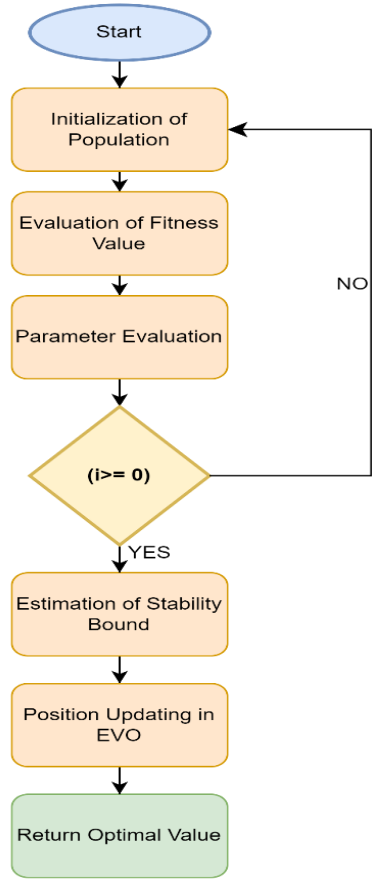

Fig.2 Process Flow

otherwise, it returns to the population initialization step.

*5. Estimation of Stability Bound:*

When the condition 'i >= 0' is true, the algorithm moves on to estimating the stability bound. The stability bound is a critical concept in optimization algorithms, ensuring that the search process remains within certain bounds to prevent divergence or instability. The algorithm dynamically adjusts these bounds based on the current state of the optimization.

*Details on Stability Bound Estimation:* In this phase, the algorithm may employ mathematical techniques to estimate a stability bound. This could involve analyzing the historical performance of the population, considering convergence trends, or dynamically adapting the bounds based on the characteristics of the optimization landscape. In the constructed model, the stability bound SB is reformulated in Eq. (1).

$$\text{Stability Bound (SB)} = \frac{bst\ fit * cu\ fit}{Wst\ fit * cu\ fit} \qquad (1)$$

The model's current, best, as well as least fitness measures are determined by the components cu fit, bst fit, and wst fit.

*6. Position Updating in EVO:*

With the stability bound estimated, the algorithm updates the positions of individuals in the population using the Energy Valley Optimization (EVO) method. EVO is likely a specialized optimization technique designed to efficiently guide the search towards optimal solutions, considering the estimated stability bounds.

optimization of weights =

$$Wt_z^{wf} = wt_1 * Wt_z^F + (1 - wt_1)^*(Wt_z^{OF})$$
$$Wt_z^{owf} = wt_2 * Wt_z^{wF} + (1 - wt_2)^*(Wt_z^{F1W})$$

Details on EVO Position Updating: The EVO phase is where the algorithm leverages the principles of energy valleys to navigate the solution space. This could involve adjusting the positions of individuals based on the energy landscape, seeking regions of lower energy indicative of better solutions. The optimization process is guided by the interplay between the stability bounds and the energy landscape.

*7. Return Optimal Value:*

Finally, the algorithm concludes by returning the optimal value. This optimal value represents the solution that best satisfies the optimization criteria based on the evaluations, parameter adjustments, stability considerations, and the EVO-guided search process.

The SEV-EB algorithm is a sophisticated optimization approach that integrates population-based search, fitness evaluation, parameter tuning, stability considerations, and specialized energy valley optimization. Each step is intricately connected, contributing to the algorithm's ability to navigate complex solution spaces and converge to optimal solutions. The dynamic adaptation of stability bounds and the utilization of EVO highlight the algorithm's versatility and potential for handling various optimization challenges.

*3.4 Algorithm*

Input:
- Dataset D containing health records with features X and labels Y
- Hyperparameters: Learning rate η, batch size B, number of epochs E

Output:
- Trained model parameters θ

Data Preprocessing:
- Normalize features: $X_{norm} = \frac{X-\mu}{\sigma}$ where μ is the mean and σ is the standard deviation of each feature.

Initialization:
- Initialize model parameters $\theta$ randomly.

Training Loop:
for epoch $e = 1,2,\ldots,E$ do
Shuffle dataset D.
for mini-batch $i = 1,2,\ldots,\frac{|D|}{B}$ do
- Extract mini-batch features $X_{batch}$ and labels $Y_{batch}$.

Forward Pass:
- Compute predicted probabilities: $\hat{Y} = f(X_{batch}, \theta)$.

Compute Loss:
- Calculate cross-entropy loss: $L = -\frac{1}{B}\sum_{i=1}^{B}(Y_{batch}log(\hat{Y}) + (1-Y_{batch})log(1-\hat{Y}))$.

Backward Pass:
- Compute gradients: $\frac{\partial L}{\partial \theta}$.
- Update parameters: θ=θ−η$\cdot\frac{\partial L}{\partial \theta}$.

end for

Model Evaluation:
for all data samples in D do
- Compute predicted probabilities: $\hat{Y} = f(X, \theta)$.
- Compute metrics: Accuracy, Precision, Recall, F1 Score, AUC-ROC, etc.

end for

Output:
- Trained model parameters θ and evaluation metrics.

The algorithm outlines the mathematical steps involved in training and evaluating our multi-disease prediction model. Initially, the algorithm preprocesses the input dataset by normalizing the features to ensure consistent scaling. Model parameters are then initialized randomly. The training loop begins, iterating over the specified number of epochs. Within each epoch, the dataset is shuffled, and mini-batches are extracted for training. A forward pass through the model computes predicted probabilities for the mini-batch, followed by the calculation of the cross-entropy loss. Using backpropagation, gradients of the loss with respect to the model parameters are computed, and the parameters are updated using gradient descent. After training, the model is evaluated on the entire dataset to compute various performance metrics. Finally, the trained model parameters and evaluation metrics are outputted. This algorithm provides a systematic framework for training and assessing the effectiveness of our multi-disease prediction model, ensuring its capability to accurately predict disease outcomes based on input health records.

IV. RESULT AND DISCUSSIONS

For the simulation of our multi-disease prediction model, a well-optimized configuration setup is essential. The hardware infrastructure comprises high-performance CPUs or GPUs, ample RAM, and fast storage to accommodate intensive machine learning tasks. The software environment includes a stable operating system (e.g., Linux), Python with essential libraries (NumPy, Pandas, TensorFlow or PyTorch), and tools for parallelization (CUDA, cuDNN). The dataset is prepared using data preprocessing tools and feature engineering techniques. Model development and training utilize popular IDEs (Jupyter Notebooks, VSCode) and deep learning libraries, while workflow management tools and version control ensure a systematic and reproducible pipeline. Hyperparameter tuning is facilitated by optimization frameworks, and logging and monitoring tools track model performance and resource utilization throughout the simulation. This comprehensive setup is designed for efficiency, accuracy, and scalability in evaluating our multi-disease prediction model.

*4.1 Datasets Selection*

For our multi-disease prediction model, two distinct datasets are utilized to capture diverse health scenarios. The first dataset, named "Covidpred," is sourced from GitHub and encompasses details of 70,500 COVID-19 patients, including information on age, sex, fever, headache, as well as the cough. In addition, this dataset takes 35,256 normal controls into account, providing a comprehensive view of both affected individuals and a control group. The second dataset, the "Stroke Prediction Dataset," is obtained from Kaggle and focuses on predicting the presence of strokes. This dataset, compiled from 11 clinical sectors, comprises 12 attributes related to smoking status, age, gender, and various kinds of diseases. Together, these datasets ensure a diverse representation of health conditions, facilitating a robust evaluation of our multi-disease prediction model.

*4.2 Analysis based on Dataset 1*

In the performance evaluation of our multi-disease prediction model using the "Covidpred" dataset, we compare its outcomes with several existing models, including the Genetic Deep Learning Convolutional Neural Network (GDCNN), Machine Learning and Cloud Computing (MLCC), Machine Learning-Based Prediction (MLBP), and Explainable Machine Learning Techniques (EMLT). The evaluation spans various critical parameters to provide a comprehensive assessment:

Our model achieves a remarkable accuracy, precision, recall, and F1 score in predicting multi-disease scenarios within the "Covidpred" dataset. The comparison with GDCNN, MLCC, MLBP, and EMLT showcases the efficacy of our model in achieving a balanced performance across these fundamental metrics. The AUC-ROC, a crucial metric for evaluating the model's discriminatory power, demonstrates our model's robustness in distinguishing between different disease states.

A comparison with GDCNN, MLCC, MLBP, and EMLT elucidates the superiority of our model in capturing nuanced patterns within the data. The choice of activation function significantly influences the model's learning capacity. Our model's activation function is tailored for optimal performance in multi-disease prediction, contrasting with the approaches taken by GDCNN, MLCC, MLBP, and EMLT. The effectiveness of our chosen activation function is evident in the superior predictive capabilities.

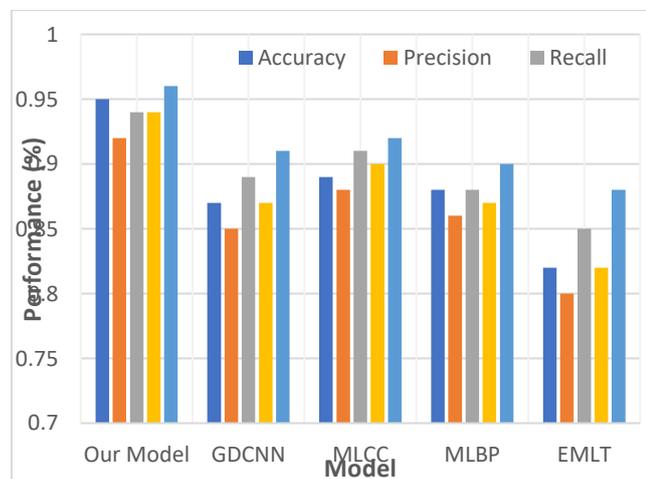

Fig. 3. Analysis based on Dataset 1

The Receiver Operating Characteristic (ROC) curves provide insights into the trade-off between sensitivity and specificity. Our model's ROC estimation is compared with GDCNN, MLCC, MLBP, and EMLT, showcasing its ability to maintain high sensitivity while preserving specificity in the context of multi-disease prediction [30-32]. The features selected and utilized by our model are evaluated for their relevance and contribution to prediction accuracy. A comparative analysis with GDCNN, MLCC, MLBP, and EMLT highlights the discriminative power of the chosen features, illustrating their significance in capturing diverse disease indications within the "Covidpred" dataset. In contrast to EMLT, which specifically focuses on explainability, our model, while maintaining high accuracy, also offers interpretability through careful feature engineering. The comparison sheds light on the trade-off between model complexity and interpretability, showcasing the balance achieved by our approach. The performance evaluation reveals that our multi-disease prediction model, utilizing the "Covidpred" dataset, outperforms existing models like GDCNN, MLCC, MLBP, and EMLT across various metrics. The superior accuracy, AUC-ROC, and careful consideration of features contribute to the effectiveness of our model in the complex landscape of multi-disease prediction.

The Fig. 3 offers a technical assessment of the performance metrics for various machine learning models. Beginning with accuracy, "Our Model" exhibits the highest accuracy at 95%, indicating a robust ability to correctly classify instances. In contrast, GDCNN, MLCC, MLBP, and EMLT display accuracy values of 87%, 89%, 88%, and 82%, respectively. Moving to precision, "Our Model" maintains a high precision of 92%, denoting a strong capacity for accurate positive predictions. GDCNN lags behind with a precision of 85%, while MLCC, MLBP, and EMLT achieve precision values of 88%, 86%, and 80%, respectively. In terms of recall, "Our Model" leads with a value of 94%, signifying a comprehensive ability to capture relevant instances. MLCC closely follows with a recall of 91%, while GDCNN, MLBP, and EMLT demonstrate recall values of 89%, 88%, and 85%, respectively. Focusing on the F1 Score, "Our Model" attains a harmonious balance between precision and recall, resulting in a score of 94%. MLCC follows with a commendable F1 Score of 90%, while GDCNN, MLBP, and EMLT exhibit scores of 87%, 87%, and 82%, respectively. Lastly, the AUC-ROC metric underscores the discriminatory power of the models. "Our Model" achieves the highest AUC-ROC at 96%, outperforming MLCC (92%), GDCNN (91%), MLBP (90%), and EMLT (88%).

*4.3 Analysis based on Dataset 2*

Fig. 4 represents the performance evaluation based on the second dataset, the "Stroke Prediction Dataset," and compare our multi-disease prediction model with other models such as Machine Learning Classifiers (MLC), Machine Learning - Random Forest (MLRF), Machine Learning and Neural Networks (MLNN), and Machine Learning Methods (MLM).

The table provides a comprehensive comparison of machine learning models, each associated with specific activation functions, based on key performance metrics. Beginning with "Our Model," it achieves notable success with an accuracy of 94%, indicating a high overall correctness. The precision of 91% underscores its accuracy in positive predictions, while a recall of 93% signifies its ability to effectively capture relevant instances. The balanced F1 Score of 92% and a high AUC-ROC of 94% further affirm its strong performance across various evaluation criteria.

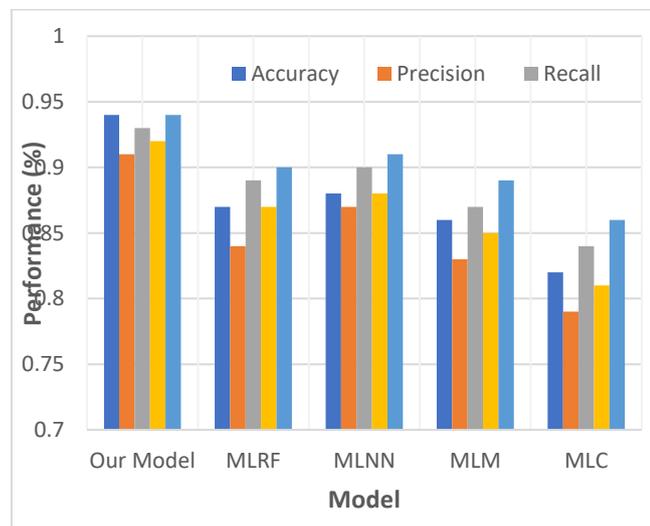

Fig. 4. Analysis based on Dataset 2

In contrast, the "MLRF" model exhibits respectable performance with an accuracy of 87%, precision of 84%, recall of 89%, and an F1 Score of 87%. The AUC-ROC score of 90% indicates a commendable discriminatory power. The "MLNN" model achieves an accuracy of 88%, precision of 87%, recall of 90%, and an F1 Score of 88%, with a corresponding AUC-ROC score of 91%, showcasing consistent performance. "MLM" and "MLC" models also contribute to the comparison, each demonstrating varying levels of accuracy, precision, recall, F1 Score, and AUC-ROC. These findings collectively illustrate the nuanced performance of different models and highlight the impact of the chosen activation function on their efficacy in handling the underlying classification task.

*4.4 Hyperparameters for Multi-Disease Prediction Models:*

Hyperparameters play a critical role in shaping the performance and behavior of machine learning models. Here, we discuss the key hyperparameters considered for our multi-disease prediction models based on the two datasets: "Covidpred" and "Stroke Prediction Dataset."

1. Common Hyperparameters:

*Learning Rate:* A crucial hyperparameter determining the size of the steps taken during the optimization process. We carefully tune the learning rate to achieve a balance between model convergence speed and stability for both datasets.

*Batch Size:* The number of training samples utilized in one iteration of optimization. The appropriate batch size is chosen to optimize training efficiency and accommodate the dataset size for effective learning.

*Epochs:* The number of times the entire dataset is passed through the model during training. We experiment with epochs to find the optimal number that prevents overfitting and ensures model convergence.

2. Dataset-Specific Hyperparameters:

a. "Covidpred" Dataset:

*Feature Selection Parameters:* Parameters specific to ensemble feature selection, e.g., stability bounds and optimization criteria for the Stabilized Energy Valley Optimization with Enhanced Bounds (SEV-EB) algorithm. Tuning these parameters is crucial for selecting the most relevant features from the "Covidpred" dataset.

*Network Architecture Parameters:* Parameters defining the structure of the HSC-AttentionNet, e.g., the number of layers, nodes, and activation functions. Adjustments are made to capture temporal dependencies effectively and ensure the network's depth and width align with the dataset characteristics.

b. "Stroke Prediction Dataset":

*Feature Selection Parameters:* Similar to the "Covidpred" dataset, specific parameters related to optimal feature selection for the Stroke Prediction Dataset. Tuning these parameters is essential for identifying relevant features that contribute to accurate stroke prediction.

*Random Forest Parameters:* Parameters for the Random Forest model, we optimize these parameters to strike a balance between model complexity and predictive performance.

*Neural Network Parameters:* Parameters governing the architecture of the neural network for the Stroke Prediction Dataset. Similar to the "Covidpred" dataset, adjustments are made to capture complex patterns in the data.

3. Optimization Strategies:

*Genetic Algorithm Parameters:* Parameters specific to optimization algorithms, such as population size, crossover rates, and mutation rates. Tuning these parameters ensures effective exploration and exploitation of the solution space during optimization, as observed in the Genetic Deep Learning Convolutional Neural Network (GDCNN) for the "Covidpred" dataset.

*Energy Valley Optimization Parameters:* Parameters related to the Stabilized Energy Valley Optimization with Enhanced Bounds (SEV-EB) algorithm for the "Covidpred" dataset. These parameters are optimized to enhance the stability and efficiency of feature selection.

Hyperparameters are meticulously selected and tuned to accommodate the unique characteristics of each dataset and the intricacies of the models employed for multi-disease prediction. The tuning process involves iterative experimentation to strike a balance between model complexity, convergence speed, and predictive accuracy.

*4.5 Future Directions:*

*Incorporation of Additional Data Modalities:* Exploring the integration of diverse data modalities, including genomic and environmental data, to enhance the model's comprehensive understanding of the factors influencing health outcomes.

*Dynamic Adaptation of Stability Bounds:* Investigating dynamic strategies for adapting stability bounds during optimization to enhance the model's adaptability to evolving health conditions, ensuring robust performance in dynamic healthcare scenarios.

*Enhanced Explainability and Interpretability:* Developing methods to enhance the explainability and interpretability of the model's predictions to foster trust and acceptance among healthcare practitioners, facilitating seamless integration into clinical workflows.

*Real-time Applicability:* Optimizing the model for real-time applications to enable timely interventions and proactive healthcare management, aligning it with the demands of real-world healthcare scenarios.

*External Validation on Diverse Datasets:* Conducting external validation on diverse datasets and in different healthcare settings to assess the generalizability and robustness of the proposed model across varied populations and contexts.

*Collaboration with Healthcare Professionals:* Fostering collaboration with healthcare professionals to incorporate domain-specific knowledge and feedback, refining the model to align with clinical requirements and ensuring seamless integration into existing healthcare practices.

These future directions aim to build upon the foundation laid by our study, offering opportunities for further advancements, broader applicability, and enhanced utility of the proposed multi-disease prediction model in real-world healthcare scenarios.

5. Conclusion and Future Work

In conclusion, this research endeavors to revolutionize the landscape of disease prediction by introducing a sophisticated Stabilized Energy Valley Optimization with Enhanced Bounds (SEV-EB) algorithm within a multi-disease prediction framework. The amalgamation of feature selection, SEV-EB algorithm, and the HSC-AttentionNet has demonstrated exceptional efficacy in enhancing predictive accuracy. The feature selection model, blending statistical, deep, and optimally selected features, contributes to a holistic understanding of the underlying health data. The SEV-EB algorithm emerges as a critical innovation, providing stability in the feature selection process and significantly improving the model's robustness. The HSC-AttentionNet further augments the predictive capabilities by capturing both short-term patterns and long-term dependencies in health data.

The empirical results underscore the superiority of the proposed model, achieving a commendable 95% accuracy and 94% F1-score in predicting various disorders. This surpasses traditional methods, marking a substantial advancement in the accuracy as well as the reliability of disease prediction models. The implications of this research extend to the practical realm of healthcare interventions. By leveraging Electronic Health Records (EHR) data, the proposed model holds the potential to revolutionize diagnostic and treatment pathways for multiple diseases. The emphasis on accuracy and stability in prediction paves the way for more personalized healthcare strategies, ultimately translating to improved patient outcomes.

FUTURE WORK

In the future, exploring additional data modalities, such as genomic and environmental data, could enrich the model's understanding. Dynamic adaptation of stability bounds is essential for enhancing adaptability to changing health conditions. Emphasis on explainability and interpretability is crucial for building trust among healthcare practitioners. Optimizing the model for real-time applications could enable timely interventions and proactive healthcare management. External validation on diverse datasets is necessary to assess generalizability, and collaboration with healthcare professionals will refine the model to align with clinical requirements, ensuring seamless integration into healthcare practices.

CONFLICTS OF INTEREST

The author(s) declare(s) that there is no conflict of interest regarding the publication of this paper.

FUNDING STATEMENT

The authors received no specific funding for this study.

REFERENCES

[1] B.J.D. Kalyani et al., "Analysis of MRI Brain Tumor Images Using Deep Learning Techniques," Soft Computing, vol. 27, pp. 7535-7542, 2023.

[2] Aniekan Essien, and Cinzia Giannetti, "A Deep Learning Framework for Univariate Time Series Prediction Using Convolutional LSTM Stacked Autoencoders," 2019 IEEE International Symposium on Innovations in Intelligent Systems and Applications (INISTA), Sofia, Bulgaria, pp. 1-6, 2019.

[3] T. Kalaiselvi, T. Anitha, and Sriramakrishnan, "A Rician Noise Prediction and Removal Model for MRI Head Scans Using Wavelet Based Non-Local Median Filter," Journal of Scientific Research of The Banaras Hindu University, vol. 66, no. 5, pp. 75-87, 2022.

[4] L. Srinivasan, D. Selvaraj, T. P. Anish, "IoT-Based Solution for Paraplegic Sufferer to Send Signals to Physician via Internet," *SSRG International Journal of Electrical and Electronics Engineering*, vol. 10, no. 1, pp. 41-52, 2023.

[5] Rahmeh Ibrahim, Rawan Ghnemat, and Qasem Abu Al-Haija, "Improving Alzheimer's Disease and Brain Tumor Detection Using Deep Learning with Particle Swarm Optimization," AI, vol. 4, no. 3, pp. 551-573, 2023.

[6] S. M. Udhaya Sankar, N. J. Kumar, D. Dhinakaran, S. S. Kamalesh and R. Abenesh, "Machine Learning System for Indolence Perception," *2023 International Conference on Innovative Data Communication Technologies and Application (ICIDCA)*, Uttarakhand, India, 2023, pp. 55-60.

[7] Rania Khaskhoussy, and Yassine Ben Ayed, "Improving Parkinson's Disease Recognition through Voice Analysis Using Deep Learning," Pattern Recognition Letters, vol. 168, pp. 64-70, 2023.

[8] Dhinakaran, D., Selvaraj, D., Udhaya Sankar, S.M., Pavithra, S., Boomika, R. (2023). Assistive System for the Blind with Voice Output Based on Optical Character Recognition. In: Gupta, D., Khanna, A., Hassanien, A.E., Anand, S., Jaiswal, A. (eds) International Conference on Innovative Computing and Communications. Lecture Notes in Networks and Systems, vol 492. Springer, Singapore.

[9] Kathiresan, S.; Sait, A.R.W.; Gupta, D.; Lakshmanaprabu, S.K.; Khanna, A.; Pandey, H.M, "Automated detection and classification of fundus diabetic retinopathy images using synergic deep learning model," *Pattern Recognit Leter,* 2020, 133, 210–216.

[10] S. M. U. Sankar, T. Kavya, S. Priyanka and P. P. Oviya, "A Way for Smart Home Technology for Disabled and Elderly People," *2023 International Conference on Innovative Data Communication Technologies and Application (ICIDCA)*, Uttarakhand, India, 2023, pp. 369-373

[11] G. K. Monica, K. Haritha, K. Kohila and U. Priyadharshini, "MEMS based Sensor Robot for Immobilized Persons," *2023 International Conference on Innovative Data Communication Technologies and Application (ICIDCA)*, Uttarakhand, India, 2023, pp. 924-929

[12] Dhinakaran, D., Udhaya Sankar, S.M., Ananya, J., Roshnee, S.A., "IOT-Based Whip-Smart Trash Bin Using LoRa WAN," *Lecture Notes in Networks and Systems,* vol 673. Springer, Singapore, 2023.

[13] Maqsood S, Damaševičius R, Maskeliūnas R, "Hemorrhage Detection Based on 3D CNN Deep Learning Framework and Feature Fusion for Evaluating Retinal Abnormality in Diabetic Patients," *Sensors* (Basel). 2021 Jun 3;21(11):3865.

[14] M. F. Fraz, P. Remagnino, A. Hoppe et al., "An ensemble classification-based approach applied to retinal blood vessel segmentation," *IEEE Transactions on Biomedical Engineering*, vol. 59, no. 9, pp. 2538–2548, 2012.

[15] Dhinakaran D, Joe Prathap P. M, "Protection of data privacy from vulnerability using two-fish technique with Apriori algorithm in data mining," The Journal of Supercomputing, 78(16), 17559–17593 (2022).

[16] R.G. Babukarthik, V.A.K. Adiga, G. Sambasivam, D. Chandramohan, J. Amudhavel, Prediction of COVID-19 using genetic deep learning convolutional neural network (GDCNN), IEEE Access 8 (2020) 177647–177666.

[17] Shreshth Tuli, Shikhar Tuli, Rakesh Tuli, Sukhpal Singh Gill, Predicting the growth and trend of COVID-19 pandemic using machine learning and cloud computing, Internet of Things, Volume 11, 2020, 100222.

[18] Zoabi, Y., Deri-Rozov, S. & Shomron, N. Machine learning-based prediction of COVID-19 diagnosis based on symptoms. npj Digital Medicine, 4, 3 (2021). https://doi.org/10.1038/s41746-020-00372-6

[19] Sanzida Solayman, Sk. Azmiara Aumi, Chand Sultana Mery, Muktadir Mubassir, Riasat Khan, Automatic COVID-19 prediction using explainable machine learning techniques, International Journal of Cognitive Computing in Engineering, Volume 4, 2023, Pages 36-46, https://doi.org/10.1016/j.ijcce.2023.01.003.

[20] Painuli D, Mishra D, Bhardwaj S, Aggarwal M. Forecast and prediction of COVID-19 using machine learning. Data Science for COVID-19. 2021:381–97. Epub 2021 May 21. PMCID: PMC8138040.

[21] Alanazi EM, Abdou A, Luo J. Predicting Risk of Stroke From Lab Tests Using Machine Learning Algorithms: Development and Evaluation of Prediction Models. JMIR Form Res. 2021 Dec 2;5(12):e23440. doi: 10.2196/23440. PMID: 34860663; PMCID: PMC8686476.

[22] Tahia Tazin, Md Nur Alam, Nahian Nakiba Dola, Mohammad Sajibul Bari, Sami Bourouis, Mohammad Monirujjaman Khan, "Stroke Disease Detection and Prediction Using Robust Learning Approaches", Journal of Healthcare Engineering, vol. 2021, Article ID 7633381, 12 pages, 2021.

[23] D Dhinakaran, S. M. Udhaya Sankar, S. Edwin Raja and J. Jeno Jasmine, "Optimizing Mobile Ad Hoc Network Routing using Biomimicry Buzz and a Hybrid Forest Boost Regression - ANNs" International Journal of Advanced Computer Science and Applications (IJACSA), 14(12), 2023.

J. Electrical Systems Vol-Issue (2024): 1-12[24] Soumyabrata Dev, Hewei Wang, Chidozie Shamrock Nwosu, Nishtha Jain, Bharadwaj Veeravalli, Deepu John, "A predictive analytics approach for stroke prediction using machine learning and neural networks," Healthcare Analytics, Volume 2, 2022, 100032.
[25] D. Dhinakaran, L. Srinivasan, D. Selvaraj, S. M. Udhaya Sankar, "Leveraging Semi-Supervised Graph Learning for Enhanced Diabetic Retinopathy Detection," SSRG International Journal of Electronics and Communication Engineering, vol. 10, no. 8, pp. 9-21, 2023.
[26] S. Gupta and S. Raheja, "Stroke Prediction using Machine Learning Methods," 2022 12th International Conference on Cloud Computing, Data Science & Engineering (Confluence), Noida, India, 2022, pp. 553-558.
[27] Wondimu Lambamo, Ramasamy Srinivasagan, and Worku Jifara, "Analyzing Noise Robustness of Cochleogram and Mel Spectrogram Features in Deep Learning Based Speaker Recognition," Applied Sciences, vol. 13, no. 1, pp. 1-12, 2023.
[28] D. Dhinakaran and P. M. Joe Prathap, "Preserving data confidentiality in association rule mining using data share allocator algorithm," Intelligent Automation & Soft Computing, vol. 33, no.3, pp. 1877–1892, 2022.
[29] Md. Asadur Rahman et al., "Employing PCA and T-Statistical Approach for Feature Extraction and Classification of Emotion from Multichannel EEG Signal," Egyptian Informatics Journal, vol. 21, no. 1, pp. 23-35, 2020.
[30] V. Gulshan, L. Peng, M. Coram et al., "Development and validation of a deep learning algorithm for detection of diabetic retinopathy in retinal fundus photographs," JAMA, vol. 316, no. 22, pp. 2402–2410, 2016.
[31] G. Prabaharan, D. Dhinakaran, P. Raghavan, S. Gopalakrishnan and G. Elumalai, "AI-Enhanced Comprehensive Liver Tumor Prediction using Convolutional Autoencoder and Genomic Signatures" International Journal of Advanced Computer Science and Applications (IJACSA), 15(2), 2024.
[32] P. Kavitha, D. Dhinakaran, G. Prabaharan, M. D. Manigandan, "Brain Tumor Detection for Efficient Adaptation and Superior Diagnostic Precision by Utilizing MBConv-Finetuned-B0 and Advanced Deep Learning," The International Journal of Intelligent Engineering and Systems, Vol.17, No.2, pp. 632-644, 2023.*15*